\pdfoutput=1

\documentclass[11pt]{article}

\usepackage[]{ACL2023}

\usepackage{times}
\usepackage{latexsym}

\usepackage[T1]{fontenc}

\usepackage[utf8]{inputenc}

\usepackage{microtype}
\usepackage{nicematrix}
\usepackage{nccmath}       
\usepackage{physics}
\usepackage{subcaption}
\usepackage[abbreviations,british]{foreign}  
\usepackage{xspace}
\usepackage{amsmath,amssymb,bm}
\usepackage[]{todonotes}
\usepackage{colortbl}
\usepackage{relsize}

\usepackage{inconsolata}
\usepackage{cleveref}

\usepackage{enumitem}

\renewcommand{\bm}[1]{\boldsymbol{\mathbf{#1}}}

\Crefformat{figure}{#2Fig.~#1#3}
\Crefmultiformat{figure}{Figs.~#2#1#3}{ and~#2#1#3}{, #2#1#3}{ and~#2#1#3}
\Crefformat{table}{#2Tab.~#1#3}
\Crefmultiformat{table}{Tabs.~#2#1#3}{ and~#2#1#3}{, #2#1#3}{ and~#2#1#3}
\Crefformat{appendix}{Appx.~\S#2#1#3}
\Crefformat{section}{\S#2#1#3}
\Crefformat{subsection}{\S#2#1#3}
\crefformat{algorithm}{Alg.~#2#1#3}
\crefformat{equation}{Eq.~#2#1#3}
\crefformat{enumi}{Prop.~#2#1#3}
\crefformat{listing}{Code.~#2#1#3}
\crefformat{lstlisting}{Code.~#2#1#3}
\Crefmultiformat{enumi}{Prop.~#2#1#3}{ and~#2#1#3}{, #2#1#3}{ and~#2#1#3}

\newcommand{\model}{\textsc{NBR}\xspace}
\newcommand{\modelmed}{\textsc{NBR}\textsubscript{NLI}\xspace}
\newcommand{\modelmednli}{\textsc{NBR}\textsubscript{NLI+FT}\xspace}
\newcommand{\modelmednliend}{\textsc{NBR}\textsubscript{NLI+FT}+EAD\xspace}
\newcommand{\ead}{\textsc{EAD}\xspace}

\newcommand{\mypar}[1]{\vspace{0.3em}\noindent\textbf{#1}}

\newcolumntype{P}[1]{>{\raggedright\arraybackslash}p{#1\textwidth-2\tabcolsep-1.5\arrayrulewidth}}
\newcolumntype{Y}[1]{>{\centering\arraybackslash}p{#1\textwidth-2\tabcolsep-1.5\arrayrulewidth}}

\newcommand{\tree}{\raisebox{5pt}{\includegraphics[scale=0.150]{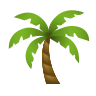}}}
\newcommand{\hugging}{\raisebox{5pt}{\includegraphics[scale=0.150]{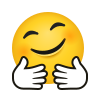}}}
\newcommand{\usc}{\raisebox{5pt}{\includegraphics[]{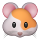}}}

\title{Can NLI Provide Proper Indirect Supervision for \\ Low-resource Biomedical Relation Extraction?}

\author{
Jiashu Xu\tree~~~
Mingyu Derek Ma\hugging~~~
Muhao Chen\usc\\
\resizebox{\textwidth}{!}{
{\tree}Harvard University~~~
{\hugging}University of California, Los Angeles~~~
{\usc}University of Southern California} \\
\texttt{jxu1@harvard.edu}~~~
\texttt{ma@cs.ucla.edu}~~~\texttt{muhaoche@usc.edu}\\
  }

\begin{document}
\maketitle

\begin{abstract}
Two key obstacles in biomedical relation extraction (RE) are the scarcity of annotations and the prevalence of instances without explicitly pre-defined labels due to low annotation coverage.
Existing approaches, which treat biomedical RE as a multi-class classification task, often result in poor generalization in low-resource settings and do not have the ability to make selective predictions on unknown cases but give a guess from seen relations, hindering the applicability of those approaches.
We present \model, which converts biomedical RE as a natural language inference formulation to provide indirect supervision. By converting relations to natural language hypotheses, \model is capable of exploiting semantic cues to alleviate annotation scarcity.
By incorporating a ranking-based loss that implicitly calibrates abstinent instances, \model learns a clearer decision boundary and is instructed to abstain on uncertain instances.
Extensive experiments on three widely-used biomedical RE benchmarks, namely ChemProt, DDI, and GAD, verify the effectiveness of \model in both full-shot and low-resource regimes.
Our analysis demonstrates that indirect supervision benefits biomedical RE even when a domain gap exists, and combining NLI knowledge with biomedical knowledge leads to the best performance gains.\footnote{Code is released at \url{https://github.com/luka-group/NLI_as_Indirect_Supervision}}
\end{abstract}

\section{Introduction}
\begin{figure*}[ht]
    \centering
    \includegraphics[
    width=0.8\textwidth, 
    ]{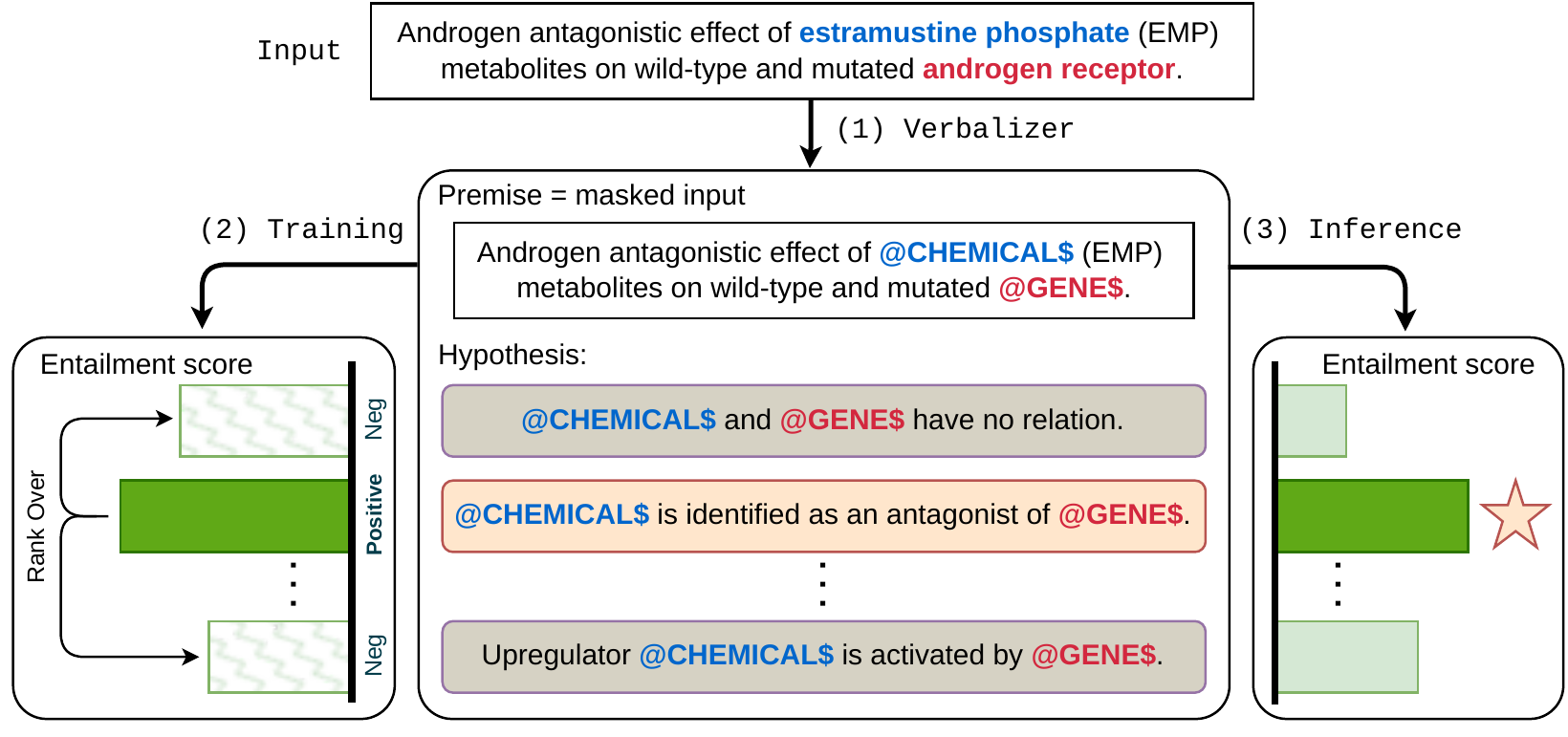}
    \caption{Overview of \model. Given an input, (1) each relation is verbalized into natural language hypotheses and masked input where entity mentions are type-masked becomes the premise.
    The ground-truth is marked in \colorbox[HTML]{FFE6CC}{light color}.
    (2) For training \model calculates the entailment scores for each relation candidate and optimizes the score of the ground-truth relation to rank over the scores of other candidates.
    (3) For inference \model computes entailment scores of each relation and returns the one with the maximum entailment score.
    }
    \label{fig:arch}
\end{figure*}

In silico studies of biology and medicine have primarily relied on machines' understanding of relations between various molecules and biomolecules.
For instance, disease-target prediction requires accurate identification of the association between the drug target and the disease \cite{bravo2015GAD}, and drug-drug interaction recognition is essential for polypharmacy side effect studies \cite{herrero2013ddi}.
Due to the complexity and high cost of human curation of such biomedical knowledge \cite{krallinger2017chemprot,bravo2015GAD}, there has been a growing interest in the field of 
biomedical relation extraction (RE), a task of automatically inferring the relations between biomedical entities described in domain-specific corpora.

However, two obstacles remain in training a reliable biomedical RE model.
First, 
biomedical RE often suffers from insufficient and imperfect annotations, due to that the annotation process is very challenging and requires expert annotators to identify complex structures from lengthy and sophisticated biomedical literature.
The existing biomedical learning resources either require very costly expert annotations \cite{krallinger2017chemprot} or resort to weak supervision \cite{bravo2015GAD}.
The insufficiency and imperfection of annotations inevitably cause existing state-of-the-art (SOTA) biomedical RE systems \cite[inter alia]
{yasunaga2022linkbert, peng2019transfer, tinn2021fine}, though showing satisfactory results in a fully supervised setting,
to result in poor generalization regarding the more common low-resource regime in this domain.
For example, \citet{han2018hierarchical} showed that model performance deteriorated quickly as the number of instances for each relation drops, 
hindering the applicability of those approaches in real-world scenarios.
Second, given that biomedical RE annotations tend to be incomplete or have low coverage, it is difficult for models to learn a clear decision boundary \cite{gardner2020evaluating}.
Specifically, in many scenarios where the described biomedical entities are not related in the context, the model may fail to abstain but give a guess from seen relations \cite{xin2021art, kamath2020selective}.
An overconfident model can be particularly harmful in high-stakes fields such as medicine, where incorrect predictions can have severe direct consequences for patients. 

Recently, indirect supervision \cite{roth2017incidental,he-etal-2021-foreseeing, levy2017zero, lu2022summarization, li2019entity} is proposed that leverages supervision signals from resource-rich source tasks to enhance resource-limited target tasks.
In this approach, the training and inference pipeline of the target task is transformed into the formulation of the source task, thus introducing additional supervision signals not accessible in the target task.
Recent works \cite{li2022ultra, yin-etal-2020-universal,sainz2021label} transfer cross-task learning signals from the Natural Language Inference (NLI) task.
The NLI task aims at 
determining whether the hypothesis can be entailed given the premise, and inductive bias of NLI models learns adaptive generalized logical reasoning  
which aligns well with the goal of biomedical RE.
On the other hand, traditional direct supervision on the biomedical RE fails to capture semantic information of relations since they are merely transformed to logits of a classifier.
By converting relations to meaningful hypotheses in NLI, the indirectly supervised method bypasses this shortage and can 
adapt the
the preexisting inductive bias of NLI-finetuned models to 
make meaningful predictions based on relation semantics \cite{huang2022unified, chen2020you}.
This critically benefits the generalizability of the model in low-resource regimes where limited direct 
supervision signals are provided \cite{sainz2021label} to 
remedy insufficient annotations.
However, previous studies focus on general domain tasks and explore little in specific domains such as biomedical.
Moreover, to maximize the utility of indirect supervision, it is found that incorporating task knowledge into the model, \ie NLI model that is trained on NLI data, yields the best performance \cite{li2022ultra, sainz2021label}.
Yet, biomedical NLI is rarely available and whether general domain NLI can provide strong indirect supervising signals to specific target domains remains unexplored.

This study presents a general learning framework, dubbed 
 \textbf{\underline{N}}LI improved \textbf{\underline{B}}iomedical \textbf{\underline{R}}elation Extraction (\model), to enhance biomedical RE with indirect supervision from \textit{general domain NLI} task. 
\Cref{fig:arch} illustrates the structure of \model.
Specifically, given an input sentence, \model reformulates RE to NLI by treating the input as the premise while verbalizing each relation label into template-based natural language hypotheses.
\model learns to 
rank the relations based on the entailment scores such that the hypothesis of a correct relation should be scored higher than those of any incorrect ones.
Furthermore, to learn a fine-grained, instance-aware decision boundary, \model deploys ranking-based loss for implicit abstention calibration that handles abstinent relations in the dataset.
During inference, the relation whose verbalized hypothesis achieved the highest score becomes the prediction.
\model fully exploits indirect supervision from NLI and performs exceptionally well even in low-resource scenarios.

Our contributions are three-fold:
First, to the best of our knowledge, this is the first work to leverage indirect supervision from NLI on biomedical RE.
Instead of solely relying on provided RE annotations, \model leverages additional supervision signals from NLI indirect supervision and can generalize well in low resource regimes.
Second, we show that \model provides a proper indirect supervision signal even if there is a domain gap between general NLI knowledge \model trained on and biomedical downstream task.
Third, we propose a new ranking-based loss that implicitly handles abstinent relations ubiquitous in biomedical RE by contrastively calibrating the score of abstinent instances.
By extensive experiments on three commonly-used biomedical RE benchmarks, namely, ChemProt \cite{krallinger2017chemprot}, DDI \cite{herrero2013ddi} and GAD \cite{bravo2015GAD}, 
we verify our contributions and show that
general domain NLI can provide a proper supervision signal, especially in low resource settings where annotations are scarce.
\model provides consistent improvements on three datasets (1.10, 1.79, and 0.96 points of F1 improvement respectively), and up to 34.25 points of F1 improvement in low-resource settings.
Further analysis demonstrates that combing NLI knowledge with biomedical knowledge leads to the best performance gains.

\section{Related Works}

\mypar{Biomedical relation extraction.}
Despite the growing availability of biomedical corpora on Web repositories, the main challenge remains in transforming
those unstructured textual data into a rigidly-structured representation that includes interested entities and relations between them \cite{peng2019transfer, lee2020biobert, tinn2021fine}.
However, knowledge curation for this purpose is often costly and requires expert involvement \cite{krallinger2017chemprot, herrero2013ddi, bravo2015GAD}.
To address this issue, biomedical RE techniques are developed to automate this process.
Most existing works
mainly conduct supervised fine-tuning 
language models pretrained on relevant corpus \eg PubMed abstracts and MIMIC-III clinical notes, on annotated biomedical RE corpora \cite{tinn2021fine, peng2019transfer, Beltagy2019SciBERT, lee2020biobert, shin-etal-2020-biomegatron, yasunaga2022linkbert}.
Two drawbacks of the aforementioned approach are: (1) it fails to capture the semantic interaction between relations and entities as relations are represented as integer indices \cite{chen2020you, huang2022unified}, and (2) performance deteriorates as the number of training instances drops \cite{han2018hierarchical}.
A straightforward solution is to curate a more expansive dataset, manually or through synthetic means \cite{schick2021generating, wu2022generating,ge2022dalle, ge2022neural, ge2023beyond, he2023annollm}.
Nonetheless, this approach entails significant costs and risk quality of the final dataset.
Consequently, we advocate for a cheaper yet efficient solution: indirect supervision.

\mypar{Indirect supervision.}
Indirect supervision \cite{roth2017incidental, he-etal-2021-foreseeing} transfers supervision signals from a more resource-rich task to enhance a specific more resource-limited task.
Often this line of work reformulates the training and inference pipeline of the target task into the form of the source task to facilitate the cross-task signal transfer.
\citet{levy2017zero} demonstrate that relation extraction can be solved using machine reading comprehension formulation. Similarly, \citet{li2019entity} and \citet{lu2022summarization} further show that relation extraction performance can be improved by multi-turn question answering and summarization, respectively.
Recently \citet{sainz2021label} and \citet{li2022ultra} propose to leverage indirect supervision from the NLI task. 
LITE (\citet{li2022ultra}) enhances entity typing by incorporating NLI and a learning-to-rank training objective while \citet{sainz2021label} observes the benefits of indirect supervision in low-resource relation extraction.
As discussed, NLI aligns well with relation extraction, but to the best of our knowledge, there is no prior work that investigates 
the effectiveness of indirect supervision when there is a domain gap between the target task and the source task, \eg biomedical domain and general domain in this study.

\section{Method}
We hereby present \model. We discuss how to frame relation extraction as a NLI task in \Cref{sub:re_NLI}, illustrate how to leverage cross-domain NLI knowledge in \Cref{sub:cross-domain-nli-finetune}, and lastly provide an optional explicit abstention detector to handle abstinent instances in \Cref{sub:end}.

\subsection{Problem Formulation}
The RE model takes a sentence $\bm{x}$ with two mentioned entities $e_1, e_2$ as input, 
and predicts the relation $y$ between $e_1, e_2$ from the label space $\mathcal{Y}$ that includes all considered relations. 
The dataset $\mathcal{D}$ consists of both non-abstinent instances where $y \in \mathcal{Y}$, and abstinent instances\footnote{
Indicating that either there is no relation between $e_1, e_2$ or the relation is not one of the relation labels defined in $\mathcal{Y}$.
} where $y=\perp$. 
A successful RE model should abstain for abstinent instances and accurately predict $y$ for non-abstinent instances.

\subsection{
Relation Extraction with NLI
}\label{sub:re_NLI}

Following \citet{sainz2021label}, we reformulate the
RE task as a NLI task, allowing cross-task transfer of indirect supervision signals from NLI resources. An overview of our pipeline is visualized in \Cref{fig:arch}.

\mypar{Decompose RE to NLI queries.}
The NLI model takes in a premise and a hypothesis, both in natural language, and outputs a logit indicating if the premise either ``entails,'' ``contradicts'' the hypothesis or the inference relation is ``neutral.'' 
We decompose an instance $(\bm{x}, e_1, e_2)$ into $|\mathcal{Y}| + 1$ NLI queries, each about a candidate relation.
We formulate the RE input sentence \bm{x} as the premise and a verbalized sentence describing the candidate relation as the hypothesis.

\mypar{Verbalizing relations to hypotheses.}
For each relation $y \in \mathcal{Y} \cup \{ \perp \}$, we verbalize $y$ as a natural language hypothesis $\nu(y)$.
Contextual textual representations of labels provide more semantic signals and are thus more understandable by a language model (LM) compared to the relation name itself or discrete relation label index used in standard classification methods \cite{chen2020you, huang2022unified}. 

Entity mentions in biomedical RE are mostly domain-specific terms that rarely appear in the LM's pre-training corpus.
The relations are always defined between entities of certain types, \eg between a gene complex and another chemical in ChemProt \cite{krallinger2017chemprot} or between two drugs in DDI \cite{herrero2013ddi}. 
Thus, each entity mention is replaced by typed entity masks such as \texttt{@GENE\$} following \citet{gu2021domain} and \citet{peng2019transfer}.\footnote{
We choose to use our typed entity mask design instead of the ``entity mask'' \cite{zhou2021improved} as it has been observed to produce better performance in those tasks with NLI.
We do not consider the entity masks as special tokens.
}
The replacement enables the LM to capture semantic information of the types and avoid using poorly trained representations for rare biomedical terms. 

As demonstrated by recent studies \cite{yeh2022decorate, li2022ultra, sainz2021label}, picking a good verbalizer for each relation may affect performance. 
Specifically, we design several types of templates (details and performances are provided in \Cref{sec:app:template_for_datasets})
listed below, each containing the two typed entity masks:
\begin{enumerate}[nolistsep,wide=\parindent,topsep=1pt]
    \item Simple Template verbalizes relation between two entities with ``\textit{is-a}'' phrase.
    \item Descriptive Template provides a contextual description of the relation.
    \item Demonstration Template includes a randomly sampled trainset exemplar with the same relation.
    \item Descriptive+Demonstration Template combines both the Descriptive description and the sampled exemplar.
    \item Learned Prompt Template \citep{yeh2022decorate} learns optimal discrete tokens for description.
\end{enumerate}
We observe that Descriptive Template performs the best 
empirically (\Cref{tab:ablations_template}).

\mypar{Confidence scoring.}
For each relation label $y \in \mathcal{Y} \cup \{\perp \}$, we calculate the confidence score of whether relation $y$ holds by $s(y) = f_{\text{NLI}}(\bm{x} \text{ [SEP] }\nu(y))$ where [SEP] is a special token separating $\bm{x}$ (premise) and $\nu(y)$ (hypothesis). $f_\text{NLI}$ is a transformer-based NLI model that encodes the input and produces logits that correspond plausibility of premise \emph{entailing} hypothesis.

\mypar{Abstention as a separate label.}
We treat $\perp$ as a separate relation label and verbalize it explicitly,
which is analogous to how supervised biomedical RE treats $\perp$ as an additional label \cite{yasunaga2022linkbert, peng2019transfer}.
An explicit template relieves the burden of incorporating both stop condition and label discriminative power into scores of $\mathcal{Y}$ labels.

\mypar{Training objective.}
\label{sub:training_objective}
Recent works in contrastive learning show that InfoNCE loss benefits efficient learning from negative examples \cite{robinson2020contrastive, wang2022simkgc, zhang2021understanding, zhou2021document, ma-etal-2023-dice, ma-etal-2021-hyperexpan-taxonomy}.
Motivated by the intuition that positive instances should be ranked higher than negative instances with regard to the anchor instance,
in each step we sample $n$ negative relations $\{y_1, \ldots, y_n\} \subseteq \mathcal{Y} \cup \{\perp\} \setminus \{y\}$ and compute $s(y_1), \ldots, s(y_n)$, and optimize ground truth relation's entailment score to be ranked higher.
Specifically, we optimize the following InfoNCE loss
\begin{align}
    \mathcal{L}_\text{NCE}  &= \sum_{(\bm{x}, y) \in \mathcal{D}} \ell_\text{NCE}(\bm{x}, y) \label{eq:NCE_loss} \\
                            &\medmath{\triangleq\!\! \sum_{(\bm{x}, y) \in \mathcal{D}} \!\!\!\!- \ln \left( \frac{\exp(s(y) / \tau)}{
            \exp(s(y)/\tau) + \sum\limits_{i=1}^{n} \exp(s(y_i) / \tau)
} \right)} \nonumber
,\end{align}

in which temperature $\tau$ controls focus on harder negatives.
In practice, learning from all possible negatives performs the best.

In pilot experiments, we observed that the model was prone to be misled by the vast number of abstinent instances in the dataset, leading to deteriorated performance. 
To alleviate such abstinent \emph{v.s.} non-abstinent imbalance, we introduce a margin-based Abstention Calibration regularization to penalize
over-confident abstinent instances while encouraging non-abstinent instances.
Concretely, if relation is not $\perp$, we calibrate the score of $\perp$ such that $s(\perp)$ is suppressed;
otherwise, we control $\perp$ to be ranked higher than other relations.
\begin{align}
    \mathcal{L}_\text{AC} &= \sum_{(\bm{x}, y) \in \mathcal{D}} \ell_\text{AC}(\bm{x}, y) \label{eq:NC_loss} \\
    \ell_\text{AC}(\bm{x}, y) &\triangleq \medmath{\begin{cases}
          \sum\limits_{i=1}^{n} \ell_\text{rank} (s(y), s(y_i); \gamma), \text{ if } y = \perp \\
          \ell_\text{rank}(s(y), s(\perp); \gamma), \text{ otherwise}
      \end{cases}} \nonumber
\end{align}
where the ranking loss $\ell_\text{rank}(x_1, x_2; \gamma)$ learns to project $x_1$ higher than $x_2$ by a margin $\gamma$.
Training with this objective, \model can be viewed as combining an implicit abstention calibrator and $s(\perp)$ as a learnable instance-aware threshold.
The final training loss is $\mathcal{L}_\text{NCE} + \lambda \mathcal{L}_\text{AC}$ where non-negative hyperparameter $\lambda$ controls the strength of abstention calibration.

\mypar{Inference.}
\model gathers hypotheses verbalized from every relation and performs ranking among the entailment scores of each hypothesis. Then the relation whose verbalized hypothesis achieves the highest score is selected as the final prediction. 

\subsection{Cross-Domain NLI Fine-tuning}
\label{sub:cross-domain-nli-finetune}
In order to maximize the benefit of NLI formulation, it is advised to use models trained on target-domain NLI dataset \cite{li2022ultra, sainz2021label}. However, available biomedical NLI training resource is limited.
As a remedy, we experiment with fine-tuning NLI models on two commonly used general domain NLI datasets, namely MNLI \cite{williams2017mnli} and SNLI \cite{bowman2015snli}, instead.
Empirically we found strong evidence (\Cref{sub:exp_results}, \Cref{sec:analysis}) that general-domain NLI knowledge can still be beneficial in the biomedical domain even if a domain gap exists.

\subsection{Explicit Abstention Detector}\label{sub:end}
Training with aforementioned $\mathcal{L}_\text{AC}$ (\Cref{eq:NC_loss}) makes \model an implicit abstention calibrator. 
As an optional post-process step, we can further improve \model by introducing an \textbf{\underline{E}}xplicit \textbf{\underline{A}}bstention \textbf{\underline{D}}ector (\ead).
This is analogous to the ``no-answer reader'' component used in previous works that detect abstinent instances explicitly \cite{back2019neurquri,hu2019read+,kundu2018nil}.

\ead is essentially another instance of \model trained separately on the same train set, but changing relation labels into binary ``has relation'' versus ``no relation'' ($\perp$).
A new verbalization template is created for ``has relation''.
For inference, we collect all differences $s_\text{\ead}(\perp) - s_\ead(\text{``has relation''})$ on the dev set. 
Then we iterate each difference as a threshold, and for one instance in the test set, \ead predicts $\perp$ only if the difference of such instance exceeds the threshold.
Once \ead is trained, \model and \ead are combined
using a simple heuristic:
resort to \model only when \ead prediction is not $\perp$ (\Cref{sec:app:end_variants}). 
In this manner, even if \ead makes a false positive prediction, since \model still retains the ability to flag $\perp$, such error can be recovered. Otherwise, we trust \ead prediction since it specializes in abstention prediction. 

\section{Experiments}\label{sec:exp}

In this section, we discuss our experiment setup (\Cref{sub:setup}) and evaluation results (\Cref{sub:exp_results}),
followed by detailed ablation studies (\Cref{sub:Ablation Study}) and analyses (\Cref{sec:analysis}).

\begin{table*}[t]
    \centering
    \begin{NiceTabular}[]{lccc}
        \textbf{Model} & \textbf{ChemProt} & \textbf{DDI} & \textbf{GAD} \\
        \hline\hline
        \Block{1-4}{\underline{\textsc{Supervised Methods}}} \\
         BioRE-Prompt$^\Diamond$ \cite{yeh2022decorate}& 67.46 & - & - \\
        BLUE-BERT$_\text{large}$ \cite{peng2019transfer} & 74.40 & 79.90 & - \\
         Sci-BERT$_\text{base}$$^\Diamond$ \cite{Beltagy2019SciBERT} & 74.93 & 81.32 \\
        Bio-BERT$_\text{base}$ \cite{lee2020biobert}& 76.46 & 80.33$^\Diamond$ & 79.83 \\
         BioMegatron \cite{shin-etal-2020-biomegatron} & 77.00 & - & - \\
         PubMed-BERT$_\text{base}$ \cite{tinn2021fine} & 77.24 & 82.36 & 82.34 \\
         Sci-Five\textsubscript{large}$^\Diamond$ \cite{phan2021scifive} & 77.48 & 82.23 & 79.21 \\
         KeBioLM \cite{yuan-etal-2021-improving} & 77.50 & 81.90 & 84.30 \\
         BioLink-BERT$_\text{base}$ \cite{yasunaga2022linkbert} & 77.57 & 82.72 & 84.39 \\
         BioM-ELECTRA$_\text{large}$ \cite{alrowili2021biom} & 78.60 & - & - \\
         BioRoBERTa$_\text{large}$ \cite{alrowili2021biom} & 78.80 & - & - \\
         BioM-ALBERT$_\text{xxlarge}$ \cite{alrowili2021biom} & 79.30 & 82.04$^\Diamond$ & - \\
         BioLink-BERT$_\text{large}$ \cite{yasunaga2022linkbert} & 79.98 & \cellcolor{cyan!25}{83.35} & \cellcolor{cyan!25}{84.90} \\
         BioM-BERT$_\text{large}$ \cite{alrowili2021biom} & \cellcolor{cyan!25}{80.00} & 81.92$^\Diamond$ & - \\
        \hline
        \Block{1-4}{\underline{\textsc{Indirect Supervision}}}\\
        \modelmed (\Cref{sub:re_NLI}) & 79.30 & 83.87 & 83.75 \\
        \modelmednli (\Cref{sub:cross-domain-nli-finetune}) & 80.54 & 84.66 & \cellcolor{red!25}{85.86} \\
         \modelmednliend (\Cref{sub:end}) & \cellcolor{red!25}{81.10} & \cellcolor{red!25}{85.14} & - \\
        \hline
         \hline
    \end{NiceTabular}
    \caption{Model performance (micro F1) using full training data on 3 biomedical RE datasets. 
    Since GAD does not contain abstinent instances, \ead is unnecessary.
    $^\Diamond$ indicates the results are from our re-implementation to conform to our evaluation metric. Other baseline performances are taken from their papers. We highlight the best results in \colorbox{red!25}{red} and the best results of direct supervision in \colorbox{cyan!25}{cyan}. 
    }
    \label{tab:full_set}
\end{table*}

\begin{table*}[ht]
\begin{subtable}[t]{1.00\textwidth}
    \centering
    \setlength{\tabcolsep}{2.5pt}
    \begin{NiceTabular}[baseline=2]{l|c|c|c|c|c|c}
        \textbf{Model on ChemProt} & \textbf{0 shot} & \textbf{8 shot} & \textbf{1\%} & \textbf{50 shot} & \textbf{10\%} & \textbf{100\%} \\
        \hline\hline
        BioRE-Prompt$^\Diamond$ \cite{yeh2022decorate} & \cellcolor{cyan!25}{1.32} & 6.07 & 27.89 & 36.80 & 55.66 & 67.46 \\
        BLUE-BERT$_\text{large}$ \cite{peng2019transfer} & - & 10.22 & 20.13 & 27.91 & 51.02 & 74.40 \\ 
        Sci-BERT$_\text{base}$$^\Diamond$ \cite{Beltagy2019SciBERT} & - & 15.60 & 22.08 & 33.36 & 60.60 & 74.93\\
        Bio-BERT$_\text{base}$ \cite{lee2020biobert} & - & 10.28 & 20.96 & 38.15 & 68.01 & 76.46 \\
        PubMed-BERT$_\text{base}$ \cite{tinn2021fine} & - & 15.97 & 23.49 & 35.37 & 68.49 & 77.24 \\
        Sci-Five\textsubscript{large}$^\Diamond$ \cite{phan2021scifive} & 0.00 & \cellcolor{cyan!25}{17.19} & \cellcolor{cyan!25}{35.66} & \cellcolor{cyan!25}{47.41} & 68.62 & 77.48 \\
        BioM-ALBERT$_\text{xxlarge}$ \cite{alrowili2021biom} & - & 8.49 & 14.95 & 21.92 & 51.69 & 79.30 \\
        BioLinkBERT$_\text{large}$ \cite{yasunaga2022linkbert} & - & 9.31 & 21.19 & 38.70 & \cellcolor{cyan!25}{71.37} & 79.98 \\
        BioM-BERT$_\text{large}$ \cite{alrowili2021biom}  & - & {16.02} & 26.23 & 40.63 & 68.93 & \cellcolor{cyan!25}{80.00} \\
        \hline
        \modelmed (\Cref{sub:re_NLI}) & 5.70 & 36.42 & 49.63 & 51.95 & 72.03 & 79.30 \\
        \modelmednli (\Cref{sub:cross-domain-nli-finetune}) & \cellcolor{red!25}{24.50} & 46.53 & 60.17 & 56.43 & 75.12 & 80.54 \\
        \modelmednliend (\Cref{sub:end}) & - & \cellcolor{red!25}{51.44} & \cellcolor{red!25}{60.34} & \cellcolor{red!25}{61.31} & \cellcolor{red!25}{75.24} & \cellcolor{red!25}{81.10}
    \end{NiceTabular}
    \label{tab:low_resource_chemprot}
\end{subtable}
\begin{subtable}[t]{1.00\textwidth}
    \centering
    \setlength{\tabcolsep}{2.5pt}
    \begin{NiceTabular}[baseline=2]{l|c|c|c|c|c|c}
        \textbf{Model on DDI} & \textbf{0 shot} & \textbf{8 shot} & \textbf{50 shot} & \textbf{1\%} & \textbf{10\%} & \textbf{100\%} \\
        \hline\hline
        BLUE-BERT$_\text{large}$ \cite{peng2019transfer} & - & 8.76 & 25.79 & 27.48 & 65.62 & 79.90 \\
        Bio-BERT$_\text{base}$ \cite{lee2020biobert} & - & 13.61 & 31.93 & 30.01 & 64.56 & 80.33 \\
        Sci-BERT$_\text{base}$$^\Diamond$ \cite{Beltagy2019SciBERT} & - & 10.55 & 33.34 & 23.62 & 69.44 & 81.32 \\
        Sci-Five\textsubscript{large}$^\Diamond$ \cite{phan2021scifive} & 0.00 & \cellcolor{cyan!25}{25.44} & \cellcolor{cyan!25}{39.36} & 29.80 & {77.11} & 82.23 \\
        PubMed-BERT$_\text{base}$ \cite{tinn2021fine} & - & 17.02 & 34.39 & 27.53 & 71.98 & 82.36 \\
        BioM-ALBERT$_\text{xxlarge}$ \cite{alrowili2021biom} & - & 11.52 & 22.50 & 18.64 & 76.70 & 82.04 \\
        BioLinkBERT$_\text{large}$ \cite{yasunaga2022linkbert} & - & 9.70 & 37.80 & \cellcolor{cyan!25}{34.11} & 74.08 & \cellcolor{cyan!25}{83.35} \\
        BioM-BERT$_\text{large}$ \cite{alrowili2021biom}  & - & 16.42 & 37.25 & 27.85 & \cellcolor{cyan!25}{79.07} & 81.92 \\
        \hline
        \modelmed (\Cref{sub:re_NLI}) & 3.60 & 32.01 & 47.86 & 53.53 & 79.49 & 83.87 \\
        \modelmednli (\Cref{sub:cross-domain-nli-finetune})  & \cellcolor{red!25}{11.94} & 37.80 & 52.49 & 60.20 & 80.85 & 84.66 \\
        \modelmednliend (\Cref{sub:end}) & - & \cellcolor{red!25}{42.48} & \cellcolor{red!25}{58.50} & \cellcolor{red!25}{61.06} & \cellcolor{red!25}{81.71} & \cellcolor{red!25}{85.14}
    \end{NiceTabular}
    \label{tab:low_resource_ddi}
\end{subtable}
\caption{
We conduct experiment on \{0,8,50\}-shot and \{1,10\}-\% ChemProt (top) and DDI (bottom).
We highlight the best model in \colorbox{red!25}{red} and the best of direct supervision in \colorbox{cyan!25}{cyan}.
Columns are ordered by the number of training instances.
$\Diamond$ indicates the results are from our re-implementation to conform to our evaluation metric. 
}
    \label{tab:low_resource}
\end{table*}

\subsection{Experimental Setup}\label{sub:setup}

\mypar{Dataset and evaluation metric.}
We conduct experiments on three sentence-level biomedical RE datasets contained in the widely-used BLURB benchmark \cite{gu2021domain}.
\textbf{ChemProt} \cite{krallinger2017chemprot} consists of PubMed abstracts corpora with five high-level chemical-protein interaction annotations.
\textbf{DDI} \cite{herrero2013ddi} studies drug-drug interaction and specializes in pharmacovigilance built from PubMed abstracts.
\textbf{GAD} \cite{bravo2015GAD} is a semi-labeled dataset created using Genetic Association Archive and consists of gene-disease associations.

There are multiple variants of the datasets used by existing literature that differ by 
data statistics or evaluation protocol
\cite{dong2021mapre, phan2021scifive, Beltagy2019SciBERT, yeh2022decorate, peng2020learning, xu2022towards} as described in \Cref{sec:app:eval_difference}, 
we adopt the most popular setting used by \citet{gu2021domain} and give dataset statistics in \Cref{tab:dataset_statistics}.
Most of entity pairs are labeled as $\perp$ without an explicit relation label.\footnote{In train set, ChemProt contains 77\% abstinent while DDI contains 85\%.}
This setting is realistic since the model must identify a relation's existence first. 
Following \citet{gu2021domain}, we use the micro F1 score calculated across all non-abstinent instances as the evaluation metric.

\mypar{Baselines.}
We compare against the various baselines (\Cref{sec:app:models}), mostly classification-based approaches that use $|\mathcal{Y}|+1$-way classification head on top of a biomedical-pretrained LM. Sci-Five \cite{phan2021scifive} generates the relation label as a seq-to-seq conditional generation formulation.

\mypar{Our method.}
We term three variants of \model:
\begin{itemize}[leftmargin=*,topsep=1pt, wide=\parindent]
\itemsep-0.2em 
    \item \modelmed 
using NLI formulation (\Cref{sub:re_NLI}) with BioLinkBERT$_\text{large}$ \cite{yasunaga2022linkbert} backbone that pretrained on biomedical corpus.
    \item \modelmednli 
further cross-domain fine-tunes (\Cref{sub:cross-domain-nli-finetune}) BioLinkBERT on two general domain NLI datasets.
The model retains biomedical domain knowledge and learns relevant NLI knowledge.
    \item \modelmednliend 
assembles \modelmednli with a separately trained \ead component (\Cref{sub:end}).
\end{itemize}

We choose BioLinkBERT as the pretrained LM due to its supremacy in performance on various biomedical domain tasks, but we emphasize that our approach is agnostic to backbone models.

\subsection{Experimental Results} \label{sub:exp_results}

\mypar{NLI provides helpful indirect supervision.}
\label{sub:NLI provide indirect supervision in medical RE}
We report the comparison between \model and baselines in \Cref{tab:full_set}.
Overall, \modelmednliend achieves SOTA performance on all three datasets, with 1.10, 1.79, and 0.96 points F1 improvement on ChemProt, DDI, and GAD respectively.
Strong performance gains verify the effectiveness of reformulating biomedical RE as NLI.
NLI supervision signals from the general domain are transferred to enhance the biomedical RE learning signals.
By verbalizing relations into natural language hypothesis, \model leverages the preexisting inductive bias of NLI-finetuned models to make informed predictions based on relation semantics. 

We further compare the performance of our model's variants. 
First, due to the prevalence of abstinent instances on the datasets, we notice that by explicitly detecting the abstinent instances, assembling \ead (\Cref{sub:end}) with \modelmednli
improves performance on ChemProt and DDI.
This is likely because explicitly detecting $\perp$ by a separate EAD model reduces the burden on \modelmednli to predict relations and identify abstinent instances at the same time.
Second, we show that cross-domain fine-tuning (\Cref{sub:cross-domain-nli-finetune}) is vital.
Compared to \modelmed, which is not trained on NLI datasets, \modelmednli resulted in significant improvements in F1 across three datasets.
This demonstrates that having prior NLI knowledge allows better utilization of the NLI formulation.
Lastly, we note that \modelmed is outperformed by its direct supervision counterpart, namely BioLinkBERT on ChemProt and GAD.
The possible reason could be that the model needs to learn to perform NLI tasks on top of the RE task without NLI training, which leads to shallower supervision signals. 
However we observe that generally, and especially in low-resource regimes, \modelmed improves over direct supervision (\Cref{sec:analysis}).

\mypar{Indirect supervision from NLI shines particularly under low-resource.}%
\label{sub:NLI benefits low resource medical RE}
We evaluate the \model under zero- and few-shot settings in \Cref{tab:low_resource}.
Following existing works \cite{peng2020learning, xu2022towards}, we train the model with 0, 8 and 50 shots and 1\% and 10\% of training instances. 
We note that classification-based methods could not adapt to the zero-shot setting.

Our experimental results show that all three variants of \model consistently achieve strong performance across all few-shot settings on all datasets, \eg 34.25 points F1 improvement on 8-shot ChemProt.
The performance of direct supervision models deteriorates dramatically as the number of training instances decreases, due to the limited learning signals.
On the contrary, \model effectively leverages indirect supervision to transform richer NLI signals to improve the RE performance.
Additionally verbalized hypotheses provide valuable semantic cues for prediction.
We also observe similar patterns as the full-set experiments: using NLI knowledge learned from NLI training data improves the performance of \modelmed, and combing \ead with \modelmednli leads to further performance gains.

Lastly, we note that as the number of training instances increases, the benefits of indirect supervision tend to decrease.
This suggests that given sufficient training signals, direct supervision can learn effectively, and the marginal returns of introducing additional NLI signals become smaller. 
In practical settings where biomedical annotations are scarce, learning with indirect supervision can lead to better performance.

\subsection{Ablation Study}%
\label{sub:Ablation Study}
\begin{table}[htpb]
    \centering
    \setlength{\tabcolsep}{4pt}
    \begin{NiceTabular}[baseline=2]{l|cc|cc}
        \Block{2-1}{\textbf{Model}} & \Block{1-2}{\textbf{ChemProt}} & & \Block{1-2}{\textbf{DDI}} & \\
        & 1\% & 100\% & 1\% & 100\% \\
        \hline\hline
        \modelmednli & \textbf{60.17} & \textbf{80.54} & \textbf{60.20} & \textbf{84.66} \\
        \hline
        \quad -$\mathcal{L}_\text{NCE}$ (\Cref{eq:NCE_loss}) & 59.63 & 79.32 & 52.50 & 83.29 \\
        \quad -$\mathcal{L}_\text{AC}$  (\Cref{eq:NC_loss}) &  57.57 & 78.68 & 50.18 & 82.94 \\
        \quad -$\mathcal{L}_\text{NCE}$-$\mathcal{L}_\text{NC}$ & 53.87 & 78.12 & 20.71 & 82.74 \\
        \hline
        MedNLI & 53.58 & 79.60 & 51.04 & 82.42
    \end{NiceTabular}
    \caption{Ablation study of \model. Micro F1 is reported for 1\% and 100\% ChemProt and DDI datasets. 
    }
    \label{tab:ablations_component}
\end{table}
\begin{figure*}[ht]
    \centering
    \includegraphics[width=0.9\textwidth]{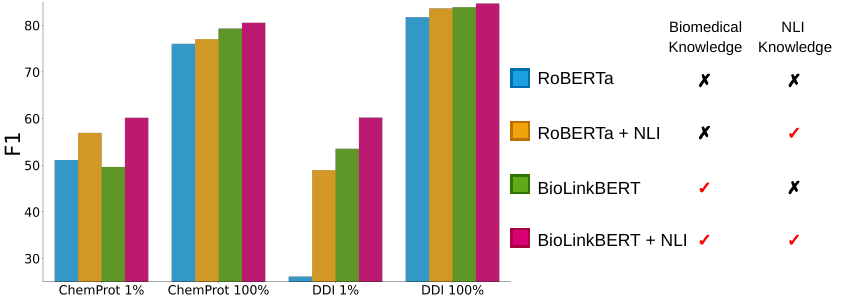}
    \caption{Impact of biomedical and NLI knowledge on 1 and 100\% ChemProt and DDI.
    Both pieces of knowledge are substantial for biomedical RE.
    }
    \label{fig:effective_IS}
\end{figure*}
We perform ablation studies on model components on ChemProt and DDI using 1\% and 100\% training data in \Cref{tab:ablations_component}.
(1) InfoNCE $\mathcal{L}_\text{NCE}$ (\Cref{eq:NCE_loss}) is essential. Replacing $\mathcal{L}_\text{NCE}$ with ranking loss sum \ie $\sum_{i=1}^n \ell_{rank}(s(y), s(y_i); \gamma)$ deteriorate performance.
These results confirm the effectiveness of InfoNCE in learning from negative samples \cite{robinson2020contrastive, wang2022simkgc}.
(2) $\mathcal{L}_\text{AC}$ (\Cref{eq:NC_loss}) is vital.
Given the prevalence of abstinent relations in the two datasets, it is easy for models to be misled by abstinent instances since they impose stronger learning signals.
We specifically notice 1\% settings have a larger performance drop, which might be caused by the fact that detecting abstention is harder
when the quantity of other labels and their associated learning signals is reduced.
(3) We further consider a variant that replaces $\mathcal{L}_\text{NCE}$ with ranking loss sum, removes $\mathcal{L}_\text{AC}$ and uses only one negative sample,
which corresponds to LITE \cite{li2022ultra} that uses NLI indirect supervision for the general domain entity typing task.
We observe further performance degradation, which again verifies the effectiveness of the two losses.
Lastly (4) we fine-tune BioLinkBERT on the biomedical MedNLI \cite{romanov2018mednli}. 
Despite being domain-relevant, we observe performance drops compared to fine-tuning on general domain NLI datasets.
We hypothesize that perform drops might be caused by (a) MedNLI being relatively small as MNLI is 35x larger 
and (b) low coverage on relevant knowledge \eg only 11.77\% of ChemProt entities are mentioned in MedNLI.
Therefore even if MedNLI provides both NLI knowledge and biomedical knowledge, the gain is insignificant.

\subsection{Analysis}
\label{sec:analysis}
In this section, we first show the benefits of indirect supervision, then illustrate two key ingredients for effective indirect supervision gains: biomedical domain knowledge and NLI knowledge.

\begin{table}[htpb]
    \centering
    \setlength{\tabcolsep}{4pt}
    \begin{NiceTabular}[baseline=2]{cc|cc|cc}
        \Block{2-2}{Dataset} & & \Block{1-4}{}\Block{1-2}{RoBERTa} & & \Block{1-2}{BioLinkBERT} & \\
         & & 
         DS & IS & DS & IS \\
        \hline\hline
        \Block{2-1}{\rotatebox{90}{\parbox{5mm}{\tiny Chem\\Prot}}} & 1\% & 0.00 & \textbf{51.11} & 21.19  & \textbf{49.63} \\
         & 100\% & 45.72 & \textbf{76.02} & \textbf{79.98} & 79.30 \\
         \hline
        \Block{2-1}{\rotate \tiny DDI} & 1\% & 15.13 & \textbf{26.11} & 34.11  & \textbf{53.53} \\
         & 100\% & 81.23 & \textbf{81.73} & 83.35 & \textbf{83.87} \\
    \end{NiceTabular}
    \caption{NLI formulation benefits, especially in low-resource settings. 
    We report performance using Direct Supervision (DS) or NLI Indirect Supervision (IS) formulation with backbones not trained on NLI datasets.}
    \label{tab:indirect_supervision_benefits}
\end{table}
\mypar{NLI formulation benefits, even without additional NLI resources.}
In \Cref{tab:indirect_supervision_benefits}, we demonstrate the effectiveness of NLI formulation using two backbones \textit{without NLI knowledge}: RoBERTa \cite{liu2019roberta} and BioLinkBERT.

We observe that 
even if models lack NLI formulation adaption, NLI formulation outperforms original RE formulation in most settings,
particularly in low-resource settings.
When data is limited, it is challenging for direct supervision methods to access sufficient supervision signals.
In contrast, the model can leverage the semantic information in the natural language hypothesis with the NLI formulation.
Additionally, BioLinkBERT consistently outperformed RoBERTa in the same settings, despite RoBERTa\textsubscript{large} having larger parameters, suggesting the importance of domain knowledge.

\mypar{Two key ingredients of indirect supervision for biomedical RE.}
We identify two potential factors that contribute to the effective usage of indirect supervision for biomedical RE:
1) biomedical domain-specific knowledge; and 2) NLI knowledge to adapt to the NLI formulation.
To test the importance of these two kinds of knowledge, in \Cref{fig:effective_IS} we evaluate on 1\% and 100\% of ChemProt and DDI the four combinations: RoBERTa and RoBERTa fine-tuned on NLI, and BioLinkBERT and BioLinkBERT fine-tuned on NLI.

We first observe that BioLinkBERT fine-tuned on NLI datasets behaves the best across all four settings, indicating the importance of both pieces of knowledge.
When the learning signal is limited, the model can dynamically load-balance both forms of knowledge to make educated predictions.
Secondly, we note that RoBERTa, which lacks both biomedical and NLI knowledge, consistently performs the worst, except for 1\% ChemProt.
Finally, it is difficult to determine whether the domain or NLI knowledge is more important in biomedical RE, as the relative importance may depend on the specific dataset or the knowledge requirements of each input.

\section{Conclusion}
We present a novel method \model that leverages indirect supervision by cross-task transfer learning from NLI tasks to improve the biomedical RE task.
\model verbalizes relations to natural language hypotheses so that model is able to exploit semantic information to make informed predictions.
Furthermore, \model adopts a ranking-based abstinent calibration loss that penalizes overconfident abstinent instances while encouraging non-abstinent instances, thus being capable of abstaining on uncertain instances.
Extensive experiments on three widely-used biomedical RE benchmarks demonstrate that \model is effective in both full-set and low-resource settings. 
We further investigate two key ingredients for effective NLI indirect supervision on biomedical RE.
Future work could involve further investigation of other indirect supervision approaches and automatic relation template generation based on prompt learning.

\section*{Acknowledgement}

We appreciate the reviewers for their insightful
comments and suggestions.
Jiashu Xu was supported by the Center for Undergraduate Research in Viterbi Engineering (CURVE) Fellowship.
Mingyu Derek Ma was supported by the AFOSR MURI grant \#FA9550-22-1-0380, the Defense Advanced Research Project Agency (DARPA) grant \#HR00112290103/HR0011260656, and a Cisco Research Award.
Muhao Chen was supported by the NSF Grant IIS 2105329, by the Air Force Research Laboratory under
agreement number FA8750-20-2-10002, by a subaward of the INFER Program through UMD ARLIS, an Amazon Research Award and a Cisco Research Award.
Computing of this work was partly supported by a subaward of NSF Cloudbank 1925001 through UCSD.

\section*{Limitations}
This work investigates using NLI as indirect supervision for biomedical RE.
Experiments suggest two key ingredients in high-performing indirect supervision biomedical RE are biomedical knowledge and NLI knowledge.
To this goal, we need to access a language model that is pretrained on biomedical domain corpus, which requires computational resources.
Compared to general domain ones, models pretrained on a specific domain are often limited in variety.
Further to learn NLI knowledge additional cross-domain fine-tuning needs to be conducted, which results in additional computational overhead.

During inference \model requires \#label times of forward passes to yield prediction since \model needs to evaluate entailment scores for each verbalized relation.
Compared to standard supervision which only requires one pass for every instance, inference cost and training cost are higher in a factor of \# label. 
Higher inference cost hinders applicability in a number of scenarios \eg real-time applications. 
Additionally, the high inference cost makes it difficult to deploy machine learning models in resource-constrained environments, such as edge devices with limited processing power.

Lastly,
since \model is sensitive to templates, designing an effective template is crucial for performance. However, currently human involvement is required to design templates for each relation.
As the number of relations increases, human involvement might become costly and time-consuming.
Moreover, it is not easy to test the effectiveness of templates as no objective metric exists, and the only way to assess the quality is to test the templates.


% ...

\bibliography{custom}
\bibliographystyle{acl_natbib}

\appendix

\begin{center}
    {
    \Large\textbf{Appendices}
    }
\end{center}

\begin{table*}[t]
    \centering
    \resizebox{\linewidth}{!}{
    \begin{NiceTabular}[baseline=2]{c|ccc|ccc|c}
        \textbf{Name} & \textbf{Relations} & \Block{1-2}{\textbf{Entity Mask}} & & \textbf{Train} & \textbf{Dev} & \textbf{Test} & \textbf{\# relations} \\
        \hline\hline
        ChemProt \cite{krallinger2017chemprot} & chemical-gene & \texttt{@CHEMICAL\$} & \texttt{@GENE\$} & 18305 & 11268 & 15745 & 5 \\
        DDI \cite{herrero2013ddi} & drug-drug & \Block{1-2}{\texttt{@DRUG\$}} & & 25296 & 2496 & 5716 & 4 \\
        GAD \cite{bravo2015GAD} & disease-gene & \texttt{@DISEASE\$} & \texttt{@GENE\$} & 4261 & 535 & 534 & 2
    \end{NiceTabular}
    }
    \caption{Dataset Statistics. \# relations does not include $\perp$. GAD does not contain abstinent instances. 
    }
    \label{tab:dataset_statistics}
\end{table*}

\section{Models}
\label{sec:app:models}
\paragraph{Baselines}
We categorize compared baselines by the pretrain corpus.
\begin{itemize}[leftmargin=*,topsep=1pt]
\itemsep-0.2em 
    \item \textit{PubMed abstracts}: \textbf{BioM-ELECTRA} \cite{alrowili2021biom}.
    
    \item \textit{PubMed abstracts and PMC full-text articles}: 
\textbf{Bio-BERT} \cite{lee2020biobert}; 
\textbf{BioM-BERT} \cite{alrowili2021biom}; 
\textbf{BioMegatron} \cite{shin-etal-2020-biomegatron} pretrain on commercial-collection subset of PMC; 
\textbf{PubMed-BERT} \cite{tinn2021fine} fine-tune model released by \citet{gu2021domain}, which is pretrain on those corpus;
\textbf{Sci-Five} \cite{phan2021scifive} is T5 based model that learns to conditionally generate relation labels in textual form directly;
\textbf{BioLinkBERT} \cite{yasunaga2022linkbert} further proposes a pretraining task of link prediction, which enables the model to learn multi-hop knowledge.

    \item \textit{PubMed abstracts and MIMIC-III clinical notes}: \textbf{BLUE-BERT} \cite{peng2019transfer}.
    
    \item \textit{Semantic Scholar}: 
\textbf{Sci-BERT} \cite{Beltagy2019SciBERT} pretrain BERT on scientific corpus consists of 1.14M full-text papers from Semantic Scholar; 
\textbf{BioRE-Prompt} \cite{yeh2022decorate} initializes from RoBERTa trained on the Semantic Scholar and learns a three-token prompt for each relation and infers by finding the best matching prompt.
\end{itemize}

We use model checkpoints released by huggingface \cite{wolf2019huggingface}.
Specifically, we use 
\texttt{bionlp/bluebert\_pubmed\_mimic\_uncased\_L\-24\_H\-1024\_A-16} for BLUE-BERT \cite{peng2019transfer}, 
\texttt{allenai/scibert\_scivocab\_uncased} for Sci-BERT \cite{Beltagy2019SciBERT},
\texttt{dmis-lab/biobert-base\-cased-v1.2} for Bio-BERT \cite{lee2020biobert}, 
\texttt{microsoft/BiomedNLP\-PubMedBERT-base-uncased-abstract-fulltext} for PubMed-BERT \cite{tinn2021fine},  
\texttt{razent/SciFive-large-Pubmed\_PMC} for Sci-Five \cite{phan2021scifive},
\texttt{sultan/BioM-ALBERT-xxlarge-PMC} for BioM-ALBERT \cite{alrowili2021biom},
\texttt{sultan/BioM-BERT-PubMed-PMC-Large} for BioM-BERT \cite{alrowili2021biom},
\texttt{michiyasunaga/BioLinkBERT-large} for BioLink-BERT \cite{yasunaga2022linkbert}, and
\texttt{cnut1648/biolinkbert-large-mnli-snli} for BioLink-BERT that is fine-tuned on SNLI \cite{bowman2015snli} and MNLI \cite{williams2017mnli}.

\paragraph{\model}
We run experiments on Quadro RTX 8000 GPU.
AdamW optimizer \cite{loshchilov2017decoupled} with learning rate 1e-5 is used, and we set margin $\gamma = 0.7$, temperature $\tau=0.01$ and calibration (\Cref{eq:NC_loss}) strength $\lambda$ in sweep from 0.001 to 10. 
We train models for 300 epochs. 
Models are evaluated every ten epochs on the dev set, and the best checkpoint is selected to infer on the test set.

\section{Evaluation Difference}
\label{sec:app:eval_difference}
As mentioned in \Cref{sec:exp}, several previous works use a different evaluation metric and variants of the datasets, rendering it hard to compare with previous work. 
In this section, we describe the main differences in the dataset. 
We first report the statistics of the dataset we use in this work in \Cref{tab:dataset_statistics}.
For other works that use variants of the datasets:
\begin{itemize}[leftmargin=*]
\itemsep-0.2em 
    \item BLUE-BERT \cite{peng2019transfer}'s variant of ChemProt and DDI. Their ChemProt contains 4,154/2,416/3458 train/val/test instances and five relations, while their DDI contains 2,937/1,004/979 train/val/test instances and four relations.
    \item Sci-BERT \cite{Beltagy2019SciBERT} uses a variant of ChemProt with 4,169/2,427/3,449 train/val/test instances and contains 13 relations.
    \item \citet{dong2021mapre} and \cite{peng2020learning} use a variant of ChemProt with 4,168/2,427/3,469 train/val/test instances and 13 relations.
    \item \citet{xu2022towards} use a variant of ChemProt with 14 relations
    \item BioRE-Prompt \cite{yeh2022decorate} also use ChemProt provided by \citet{gu2021domain}, but does not exclude abstinent instances.
\end{itemize}

\section{\ead Details and Variants}
\label{sec:app:end_variants}
\begin{table}[htpb]
    \centering
    \small
    \begin{NiceTabular}[baseline=2]{cc}
        \textbf{Heuristic} & \textbf{ChemProt} \\
        \hline\hline
        Simple & \textbf{81.10} \\
        Voting & 80.73 \\
        Confident & 80.96 \\
        Super-confident & 80.66 \\
        Classification & 80.78 \\
    \end{NiceTabular}
    \caption{
        \modelmednliend performance on ChemProt under various heuristics.
    }
    \label{tab:ead_heuristic}
\end{table}
Since only relations for \ead is ``has relation'' versus ``no relation'', instead of \Cref{eq:NCE_loss} and \Cref{eq:NC_loss} used in \model, \ead learns only via ranking loss $\ell_\text{rank}(s(y), s(y'); \gamma)$ where $y$ is the ground-truth while $y'$ is the opposite relation.

We discuss several heuristics in assembling \model and \ead.
The best performing heuristic is simple: only resort to \model when \ead prediction is not $\perp$. In other words, the final prediction is $\perp$ only if \ead prediction is $\perp$; otherwise, return the prediction of $\model$. 
We evaluate other more sophisticated heuristics:
\begin{itemize}[leftmargin=*]
\itemsep-0.2em 
    \item Voting: Predict $\perp$ only when both \model and \ead predict $\perp$; 
    otherwise, return \model's prediction.
    \item Confident: Predict $\perp$ only when \ead predicts $\perp$ and confidence score $s_\text{\ead}(\perp)$ is higher than confidence score $s_\text{\model}(\perp)$; 
    otherwise, return \model's prediction.
    Note that if \ead makes a false positive, \model is still able to recover if $s_\text{\model}(\perp)$ is the highest.
    \item Super-confident:  Predict $\perp$ when \ead predicts $\perp$; if $s_\text{\ead}(\perp) > s_\text{\model}(\perp)$ return highest-scored non-abstinent relation $\arg\max_{y \in \mathcal{Y}} s_\text{\model}(y)$; otherwise prediction of \model.
    \item Classification: Use a classification-based model (with the same backbone as \modelmednli), and use logits for confidence score under the simple heuristic.
\end{itemize}
In \Cref{tab:ead_heuristic}, we observe that a more complicated heuristic does not entail better performance gains.
Note that designing a contextual description for ``has relation'' is challenging and our template is a simple phrase such as ``relation exists between.''
Surprisingly, we still found assembling 
\model with \ead empirically outperforms classification-based 
abstention detector.
We credit enhanced performance to additional semantic information captured by the verbalized template. 

\begin{table}[htpb]
    \centering
    \resizebox{\linewidth}{!}{
    \begin{NiceTabular}[baseline=2]{l|cc|cc}
        \Block{2-1}{\textbf{Template}} & \Block{1-2}{\textbf{ChemProt}} & & \Block{1-2}{\textbf{DDI}} & \\
        & 1\% & 100\% & 1\% & 100\% \\
        \hline\hline
        Descriptive & 60.17 & \textbf{80.54} & \textbf{60.20} & \textbf{84.66} \\
        Simple & \textbf{63.80} & 79.84 & 55.38 & 83.26 \\
        Demonstration &  48.72 & 79.88 & 45.81 & 83.46 \\
        Descriptive + Demonstration  & 53.39 & 79.79 & 49.78 & 83.45 \\
        Learned Prompt & 59.45 & 79.74 & - & - \\
    \end{NiceTabular}
    }
    \caption{Ablation study of \modelmednli using different templates. Micro F1 is reported.
    \citet{yeh2022decorate} only reports results on ChemProt.
}
    \label{tab:ablations_template}
\end{table}
\section{Template for datasets}
\label{sec:app:template_for_datasets}
We provide details for each of the templates investigated in this work.
\begin{enumerate}
    \item Simple Template:
    This template verbalizes the relation between two entities as a ``\textit{is-a}'' phrase, \eg ``@CHEMICAL\$ \textit{is a downregulator} to @GENE\$.''

    \item Descriptive Template:
    We manually curate a description for each relation that contains more context, \eg ``Downregulator @CHEMICAL\$ is designed as an inhibitor of @GENE\$.''
    
    \item Demonstration Template:
    Motivated by few-shot exemplars used for in-context learning, the demonstration template includes a randomly sampled context sentence whose entities hold the same relation, \eg ``Relation described between @CHEMICAL\$ to @GENE\$ is similar to <\textit{example sentence}>.''
    
    \item Descriptive + Demonstration: 
    We include both a contextual description and an in-context exemplar by simple concatenating. 
    
    \item Learned Prompt Template:
    Borrowed from \citet{yeh2022decorate}, which leverage prompt tuning with rules \cite{han2021ptr} to learn optimal discrete tokens to fill in [MASK] within the template such as ``@CHEMICAL\$ [MASK] [MASK] [MASK] @GENE\$.''
\end{enumerate}
We further provide templates for \model on three datasets: ChemProt (\Cref{tab:dataset_templates_chemprot}),
DDI (\Cref{tab:dataset_templates_ddi}) and 
GAD (\Cref{tab:dataset_templates_gad}).
\begin{table*}[ht]
    \centering
    \small
    \begin{NiceTabular}[baseline=2]{c|P{0.5}}
        \textbf{Relation} & \textbf{Verbalized Hypothesis} \\
        \hline\hline
        0 & There is no relation between @GENE\$ and @DISEASE\$. \\
        1 & @GENE\$ and @DISEASE\$ are correlated. \\
        \hline
    \end{NiceTabular}%
    \caption{Descriptive templates on GAD.}
    \label{tab:dataset_templates_gad}
\end{table*}

\begin{table*}[ht]
    \centering
    \resizebox*{0.9\textwidth}{!}{
    \begin{NiceTabular}[baseline=2]{cc|P{0.9}}
        \Block{1-2}{\textbf{Relation}} & & \textbf{Verbalized Hypothesis} \\
        \hline\hline
        & 0 (no relation) & @DRUG\$ and @DRUG\$ are not interacting. \\
        \hline
        \Block{4-1}{\rotate Simple} & DDI-advise & Interaction described bewteen two @DRUG\$ and @DRUG\$ is about advise. \\
        & DDI-effect & Interaction described bewteen two @DRUG\$ and @DRUG\$ is about effect.\\
        & DDI-int & Interaction described bewteen two @DRUG\$ and @DRUG\$ might or maybe occur.\\
        & DDI-mechanism & Interaction described bewteen two @DRUG\$ and @DRUG\$ is about mechanism. \\
        \hline
        \Block{4-1}{\rotate Descriptive}  &  DDI-advise & A recommendation or advice regarding two @DRUG\$ is described.\\
        & DDI-effect & Medical effect regarding two @DRUG\$ is described.\\
        & DDI-int & Interaction regarding two @DRUG\$ might or maybe occur.\\
        & DDI-mechanism & Pharmacokinetic mechanism regarding two @DRUG\$ is described.\\
        \hline
        \Block{4-1}{\rotate Demonstration} & DDI-advise & The interaction between two @DRUG\$ is the same as \textcolor{cyan}{\it``perhexiline hydrogen maleate or @DRUG\$ (with hepatotoxic potential) must not be administered together with @DRUG\$ or Bezalip retard.''} \\
        & DDI-effect & The interaction between two @DRUG\$ is the same as \textcolor{cyan}{\it ``@DRUG\$ administered concurrently with @DRUG\$ reduced the urine volume in 4 healthy volunteers.''} \\
        & DDI-int & Interaction between two @DRUG\$ is the same as \textcolor{cyan}{\it @DRUG\$ may interact with @DRUG\$, butyrophenones, and certain other agents.''} \\
        & DDI-mechanism & The interaction between two @DRUG\$ is the same as \textcolor{cyan}{\it @DRUG\$, enflurane, and halothane decrease the ED50 of @DRUG\$ by 30\% to 45\%.''}\\
        \hline
        \Block{4-1}{\rotate Descriptive + Demonstration} & DDI-advise & A recommendation or advice regarding two @DRUG\$ is described, similar to \textcolor{cyan}{\it ``perhexiline hydrogen maleate or @DRUG\$ (with hepatotoxic potential) must not be administered together with @DRUG\$ or Bezalip retard.''} \\
        & DDI-effect & Medical effect regarding two @DRUG\$ is described, similar to \textcolor{cyan}{\it''@DRUG$ administered concurrently with @DRUG$ reduced the urine volume in 4 healthy volunteers.''} \\
        & DDI-int & Interaction regarding two @DRUG\$ might or maybe occur, similar to \textcolor{cyan}{\it @DRUG\$ may interact with @DRUG\$, butyrophenones, and certain other agents.''}\\
        & DDI-mechanism & Pharmacokinetic mechanism regarding two @DRUG\$ is described, similar to \textcolor{cyan}{\it ``@DRUG\$, enflurane, and halothane decrease the ED50 of @DRUG\$ by 30\% to 45\%.''} \\
        \hline
    \end{NiceTabular}%
    }
    \caption{Each variant of templates on DDI. \textcolor{cyan}{Cyan sentence} is an example from the train set.}
    \label{tab:dataset_templates_ddi}
\end{table*}

\begin{table*}[ht]
    \centering
    \begin{NiceTabular}[baseline=2]{cc|P{0.83}}
        \Block{1-2}{\textbf{Relation}} & & \textbf{Verbalized Hypothesis} \\
        \hline\hline
        & 0 (no relation) & @CHEMICAL\$ and @GENE\$ have no relation. \\
        \hline
        \Block{5-1}{\rotate Simple} & CPR:3 & @CHEMICAL\$ is a upregulator to @GENE\$.\\
        & CPR:4 & @CHEMICAL\$ is a downregulator to @GENE\$. \\
        & CPR:5 & @CHEMICAL\$ is a agonist to @GENE\$. \\
        & CPR:6 & @CHEMICAL\$ is a antagonist to @GENE\$. \\
        & CPR:9 & @CHEMICAL\$ is a substrate to @GENE\$. \\
        \hline
        \Block{5-1}{\rotate Descriptive}  & CPR:3 & Upregulator @CHEMICAL\$ is activated by @GENE\$.\\
        & CPR:4 & Downregulator @CHEMICAL\$ is designed as an inhibitor of @GENE\$. \\
        & CPR:5 & Activity of agonist @CHEMICAL\$ is mediated by @GENE\$. \\
        & CPR:6 & @CHEMICAL\$ is identified as an antagonist of @GENE\$. \\
        & CPR:9 & @CHEMICAL\$ is a substrate for @GENE\$. \\
        \hline
        \Block{5-1}{\rotate Demonstration} & CPR:3 & Relation of @CHEMICAL\$ to @GENE\$ is similar to relation described in \textcolor{cyan}{\it ``@CHEMICAL\$ selectively induced @GENE\$ in four studied HCC cell lines.''} \\
        & CPR:4 & Relation of @CHEMICAL\$ to @GENE\$ is similar to relation described in \textcolor{cyan}{\it ``@CHEMICAL\$, a new @GENE\$ inhibitor for the management of obesity.''}\\
        & CPR:5 & Relation of @CHEMICAL\$ to @GENE\$ is similar to relation described in \textcolor{cyan}{\it ``Pharmacology of @CHEMICAL\$, a selective @GENE\$/MT2 receptor agonist: a novel therapeutic drug for sleep disorders.''} \\
        & CPR:6 & Relation of @CHEMICAL\$ to @GENE\$ is similar to relation described in \textcolor{cyan}{\it ``@CHEMICAL\$ is an @GENE\$ antagonist that is metabolized primarily by glucuronidation but also undergoes oxidative metabolism by CYP3A4.''} \\
        & CPR:9 & Relation of @CHEMICAL\$ to @GENE\$ is similar to relation described in \textcolor{cyan}{\it ``For determination of [@GENE\$+Pli]-activity, @CHEMICAL\$ was added after this incubation.''} \\
        \hline
        \Block{5-1}{\rotate Descriptive + Demonstration} & CPR:3 & Upregulator @CHEMICAL\$ is activated by @GENE\$, similar to relation described in \textcolor{cyan}{\it ``@CHEMICAL\$ selectively induced @GENE\$ in four studied HCC cell lines.''} \\
        & CPR:4 & Downregulator @CHEMICAL\$ is designed as an inhibitor of @GENE\$, similar to relation described in \textcolor{cyan}{\it ``@CHEMICAL\$, a new @GENE\$ inhibitor for the management of obesity.''} \\
        & CPR:5 & Activity of agonist @CHEMICAL\$ is mediated by @GENE\$, similar to relation described in \textcolor{cyan}{\it ``Pharmacology of @CHEMICAL\$, a selective @GENE\$/MT2 receptor agonist: a novel therapeutic drug for sleep disorders.''} \\
        & CPR:6 & @CHEMICAL\$ is identified as an antagonist of @GENE\$, similar to relation described in \textcolor{cyan}{\it ``@CHEMICAL\$ is an @GENE\$ antagonist that is metabolized primarily by glucuronidation but also undergoes oxidative metabolism by CYP3A4.''} \\
        & CPR:9 & CHEMICAL\$ is a substrate for @GENE\$, similar to relation described in \textcolor{cyan}{\it ``For determination of [@GENE\$+Pli]-activity, @CHEMICAL\$ was added after this incubation.''} \\
        \hline
        \Block{5-1}{\rotate Learned Propmt} & CPR:3 & @CHEMICAL\$ is activated by @GENE\$.\\
        & CPR:4 & @CHEMICAL\$ activity inhibited by @GENE\$.\\
        & CPR:5 & @CHEMICAL\$ agonist actions of @GENE\$.\\
        & CPR:6 & @CHEMICAL\$ identified are antagonists @GENE\$.\\
        & CPR:9 & @CHEMICAL\$ is substrate for @GENE\$. \\
        \hline
    \end{NiceTabular}%
    \caption{Each variant of templates on ChemProt. \textcolor{cyan}{Cyan sentence} is an example from the train set.}
    \label{tab:dataset_templates_chemprot}
\end{table*}

Lastly, \Cref{tab:ablations_template} shows the effect of template design.
The descriptive template, which involves manual efforts, leads to the best performance.
The simple template preserves the relation name semantics and yields strong performance.
On the other hand, while popular in in-context learning works, we find that the demonstration template or descriptive + demonstration template consistently underperforms the descriptive template, indicating that incorporating examples in NLI hypothesis is not helpful potentially due to limited diversity.
The learned prompt template used by \citet{yeh2022decorate} does not outperform the manually constructed descriptive template.
Finally, we note that changing templates can lead to significant performance perturbations, 
our experiments suggest that evaluating the quality of templates in low-resource settings such as 1\% can be effective and efficient.
We note that the contextual template might not be optimal and we leave how to automatically pick the optimal template as future work.

\end{document}